\documentclass[sigconf]{acmart}

\AtBeginDocument{%
  \providecommand\BibTeX{{%
    \normalfont B\kern-0.5em{\scshape i\kern-0.25em b}\kern-0.8em\TeX}}}

\copyrightyear{2024}
\acmYear{2024}
\setcopyright{acmlicensed}\acmConference[SIGIR '24]{Proceedings of the 47th International ACM SIGIR Conference on Research and Development in Information Retrieval}{July 14--18, 2024}{Washington, DC, USA}
\acmBooktitle{Proceedings of the 47th International ACM SIGIR Conference on Research and Development in Information Retrieval (SIGIR '24), July 14--18, 2024, Washington, DC, USA}
\acmDOI{10.1145/3626772.3657855}
\acmISBN{979-8-4007-0431-4/24/07}
\begin{document}
	
\title{TriviaHG: A Dataset for Automatic Hint Generation from Factoid Questions}

\author{Jamshid Mozafari}
\orcid{0000-0003-4850-9239}
\affiliation{%
  \institution{University of Innsbruck}
  \city{Innsbruck}
  \state{Tyrol}
  \country{Austria}}
\email{jamshid.mozafari@uibk.ac.at}

\author{Anubhav Jangra}
\orcid{0000-0001-5571-6098}
\affiliation{%
  \institution{Columbia University}
  \city{New York}
  \state{New York State}
  \country{USA}}
\email{anubhav@cs.columbia.edu}

\author{Adam Jatowt}
\orcid{0000-0001-7235-0665}
\affiliation{%
  \institution{University of Innsbruck}
  \city{Innsbruck}
  \state{Tyrol}
  \country{Austria}}
\email{adam.jatowt@uibk.ac.at}

%% \renewcommand{\shortauthors}{Jamshid and Adam}

\begin{abstract}
Nowadays, individuals tend to engage in dialogues with Large Language Models, seeking answers to their questions. In times when such answers are readily accessible to anyone, the stimulation and preservation of human's cognitive abilities, as well as the assurance of maintaining good reasoning skills by humans becomes crucial. This study addresses such needs by proposing hints (instead of final answers or before giving answers) as a viable solution. We introduce a framework for the automatic hint generation for factoid questions, employing it to construct TriviaHG, a novel large-scale dataset featuring 160,230 hints corresponding to 16,645 questions from the TriviaQA dataset. Additionally, we present an automatic evaluation method that measures the Convergence and Familiarity quality attributes of hints. To evaluate the TriviaHG dataset and the proposed evaluation method, we enlisted 10 individuals to annotate 2,791 hints and tasked 6 humans with answering questions using the provided hints. The effectiveness of hints varied, with success rates of $96\%$, $78\%$, and $36\%$ for questions with easy, medium, and hard answers, respectively. Moreover, the proposed automatic evaluation methods showed a robust correlation with annotators' results. Conclusively, the findings highlight three key insights: the facilitative role of hints in resolving unknown questions, the dependence of hint quality on answer difficulty, and the feasibility of employing automatic evaluation methods for hint assessment.
\end{abstract}

\begin{CCSXML}
<ccs2012>
   <concept>
       <concept_id>10002951.10003317.10003338</concept_id>
       <concept_desc>Information systems~Retrieval models and ranking</concept_desc>
       <concept_significance>500</concept_significance>
       </concept>
   <concept>
       <concept_id>10002951.10003317.10003359</concept_id>
       <concept_desc>Information systems~Evaluation of retrieval results</concept_desc>
       <concept_significance>500</concept_significance>
       </concept>
 </ccs2012>
\end{CCSXML}

\ccsdesc[500]{Information systems~Retrieval models and ranking}
\ccsdesc[500]{Information systems~Evaluation of retrieval results}

\keywords{Hint Generation, Question Answering, Large Language Models}



\maketitle

\section{Introduction}\label{s:introduction}
Automatic question answering (QA) systems~\cite{Abdel-Nabi2023, 10.1145/3560260,mavi2022survey} have recently gained significant attention, allowing users to pose questions and receive direct answers. Traditional QA systems excel in specific knowledge domains by extracting relevant text spans from passages. In contrast, Large Language Models (LLMs)~\cite{team2023gemini, workshop2022bloom, NEURIPS2020_1457c0d6}, 
pre-trained on vast datasets, offer a more versatile approach. LLMs can handle multiple NLP tasks, including automatic QA, and can answer questions across diverse domains. Unlike extractive QA systems, they generate answers by creating new content, showcasing high proficiency and potential as an alternative to traditional QA systems.

\begin{figure}[]
  \centering
  \includegraphics[width=\linewidth]{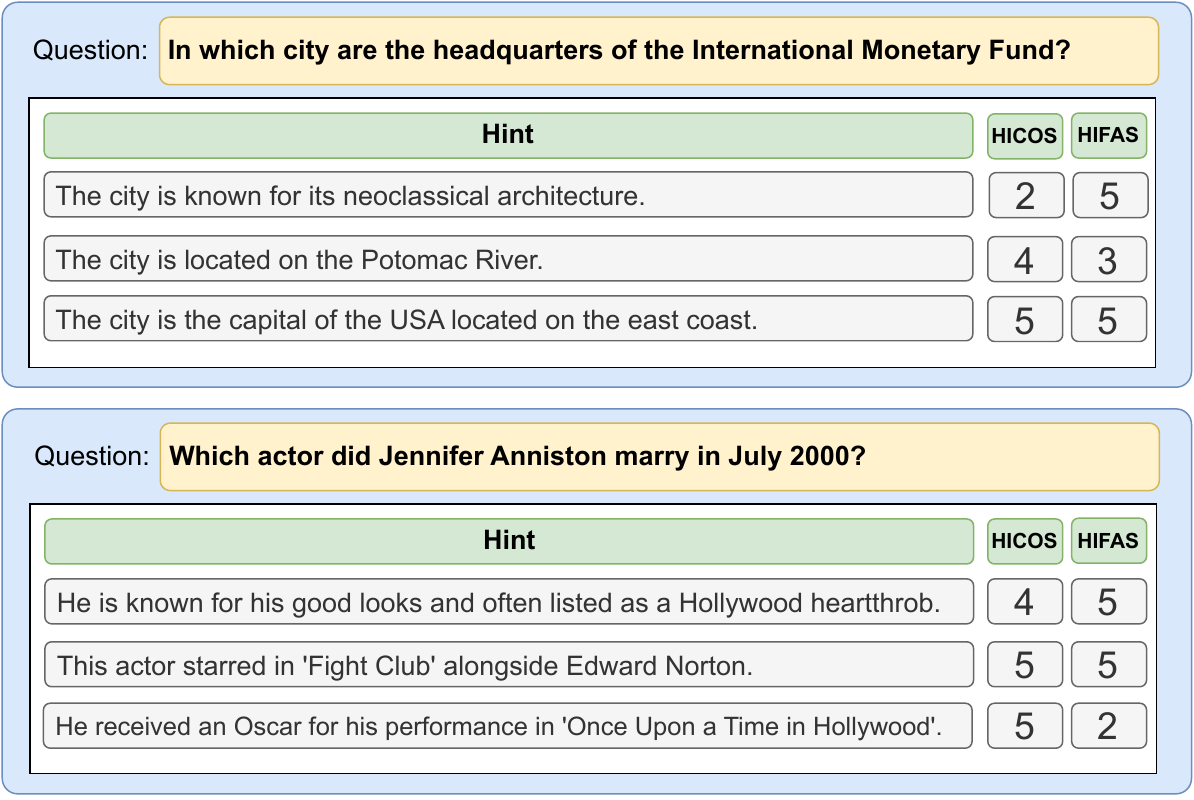}
  \caption{Example hints for two sample questions in TriviaHG with their computed convergence quality (HICOS) and familiarity (HIFAS) provided on a scale of 1 to 5 (1 is the lowest and 5 is the highest quality).}
  \label{fig:dataset_instance}
\end{figure}

While advanced systems like LLMs excel in answering questions, they raise concerns about potentially contributing to weakening essential human skills like deep thinking, reasoning, critical analysis, and even self-confidence~\cite{bandura2013role}. The problem lies in that advanced AI systems like LLMs discourage users from independently contemplating and solving questions, hindering the development of users' reasoning, memorization, and analytical abilities. Since the answers to user questions can be readily given at any time and are usually correct, then human activities that are common in traditional information searching and browsing are becoming less necessary and may occur with lesser intensity. We note here that although the current LLMs are obviously not perfect (hallucinations, inability to answer complex questions, etc.), we expect the models to keep improving in the future (judging from the recent speed of technological advancement).
Furthermore, psychological and cognitive studies highlight the importance of allowing individuals to derive answers independently, fostering self-confidence, and promoting further learning~\cite{USHER2006125}. Hence, occasionally letting users come up with answers to their questions themselves through suitable guidance should have a positive psychological effect and encourage self-learning. Finally, helping to answer questions is quite common in a variety of quizzes and quiz games, which are enjoyed by many.

In view of the above, the approaches that involve humans in the answering process instead of relying solely on answers generated by automated systems should become useful, perhaps not for everyone, but at least for interested users. We then focus in this work on using LLMs to provide hints for questions rather than generating definitive answers. Hints are clues that guide individuals towards a solution without explicitly providing the answer~\cite{doi:10.1207/s15327809jls0501_2}. They promote thinking, reasoning, and memorization skills, as humans must engage in active and thoughtful consideration after reading the hints instead of passively receiving ready answers. Additionally, relying on human input might enhance the reliability and trust in the answers as individuals draw on their own knowledge.

The use of hints to assist users is not a new concept in computer science. For example, in the field of automatic hint generation for programming (AHGP) ~\cite{10.1145/3469885}, the task is to generate hints automatically to aid students in coding exercises. However, there is limited research on automatic hint generation specifically for factoid questions (AHGQ) - questions that can be answered with facts expressed in a few words~\cite{ 10.1145/3578337.3605119}. While both AHGP and AHGQ aim to assist humans, the former is more tailored to coding exercises, often suggesting code edits, and the latter is a more general task applicable across various types of questions. Furthermore, there is an additional benefit from focusing on AHGQ task. Hints for questions could be used to enhance Retrieval Augmented Generation (RAG) solutions~\cite{gao2023retrieval, abdallah2023generator} and Query Expansion~\cite{wang-etal-2023-query2doc} methods. Specifically, in the context of Retrieval Augmented Generation, incorporating hints as query explanations has the potential to enhance the model's ability to understand and process queries.As mentioned above, the hints may also find applications in entertainment, such as in game shows like Jeopardy.

As a novel task, AHGQ faces challenges such as the absence of datasets and reliable evaluation methods. Addressing this gap, we introduce a framework for AHGQ and utilize it to generate TriviaHG\footnote{The code, dataset, and experimental results are freely available at \url{https://github.com/DataScienceUIBK/TriviaHG}}, a dataset for automatic hint generation for factoid questions. Our approach relies on web pages as the primary information source and utilizes Copilot (formerly known as Bing Chat AI)\footnote{\url{https://copilot.microsoft.com}} to generate hints based on online sources, ensuring accuracy and relevance. The hints are designed to be user-friendly and helpful.

As a second contribution, we propose an automatic evaluation method to assess hint quality, considering aspects of convergence and familiarity. The suggested evaluation method is versatile, suitable for application across various types of hints, regardless of their format, whether they be in documents, keywords, or other forms. The convergence quality is a characteristic that reflects the extent to which a hint can narrow down or eliminate potential answers to a given question. The familiarity is a characteristic that signifies the degree to which the entities mentioned in a hint are well-known or widely recognized. We label the score of convergence quality as \textit{HICOS} (\textit{HInt COnvergence Score}), and the score of familiarity quality as \textit{HIFAS} (\textit{HInt FAmiliarity Score}). Figure~\ref{fig:dataset_instance} illustrates two questions along with a few examples hints from TriviaHG along with the computed HICOS and HIFAS.

To sum up, we make the following contributions in this work:
\begin{itemize}
    \item We release a dataset called TriviaHG for automatic hint generation for questions, which includes 16,645 questions and 160,230 hints about persons, locations, entities, dates, and numbers.
    \item We propose an automatic evaluation method for assessing the quality of convergence and familiarity of hints for AHGQ.
    \item The analysis of the TriviaHG dataset and the automatic evaluation method demonstrates their quality and effectiveness, validating the framework and evaluation approach employed in the study.

\end{itemize}

\section{Related Works}\label{s:related_works}

\subsection{QA Task}\label{ss:related_qa}
QA systems can be divided into two categories: Extractive QA and Generative QA. Extractive QA systems~\cite{9991478} try to extract the answer from the passage. \citet{ramnath-etal-2020-towards} fine-tune the BERT~\cite{devlin-etal-2019-bert} language model on QA datasets and employ the fine-tuned model to extract the final answer from the given passage. In the same way, \citet{9653574} utilize RoBERTa~\cite{liu2019roberta} language model instead of BERT to extract the underlying semantic features of inputs more effectively. Generative QA systems~\cite{chen2023benchmarking}, on the other hand, try to generate the final answer. Generative QA, as opposed to extractive QA, offers advantages such as the ability to combine information from multiple sources to create coherent responses and generate original content, making it more flexible in language and capable of reasoning by comparing facts or drawing conclusions. \citet{yu2023generate} propose employing generative models such as LLMs to generate contextual documents rather than retrieving the documents from a collection. They show that their method has better results on TriviaQA~\cite{joshi-etal-2017-triviaqa} and WebQA~\cite{berant-etal-2013-semantic} datasets. \citet{izacard-grave-2021-leveraging} leverage passage retrieval component along with generative models for generating final answer using a pool of retrieved documents.

Multiple QA datasets have been proposed in various domains, such as Legal~\cite{Zhong_Xiao_Tu_Zhang_Liu_Sun_2020}, Medical~\cite{pmlr-v174-pal22a}, News~\cite{10.1145/3477495.3531734}, etc. In these datasets, each record includes three main parts: a question, an answer, and a passage such that the answer is a span of the passage. SQuAD~\cite{1360574093551088000} consists of 100,000+ questions posed by crowd-workers on 536 Wikipedia articles. Natural Questions~\cite{kwiatkowski-etal-2019-natural} contains 300,000 questions and is generated by sampling questions issued to the Google search engine. TriviaQA~\cite{joshi-etal-2017-triviaqa}, which we use in our research, is an extensive open-domain QA dataset sourced from trivia games, quizzes, and open-domain question collections, covering a broad spectrum of topics. With over 98,000 question-answer pairs and diverse linguistic forms, TriviaQA serves as a robust benchmark challenging systems to excel in comprehension and reasoning across various domains, driving advancements in machine reading comprehension. Also, some works utilize LLMs to generate datasets~\cite{liu2023secqa, 10.1145/3477495.3531734, mallen-etal-2023-trust}. SecQA~\cite{liu2023secqa} is a dataset containing 242 questions generated by GPT-4~\cite{achiam2023gpt} based on the \textit{Computer Systems Security: Planning for Success} textbook~\cite{tolboom2023computer}. 

The prior efforts have not focused on generating hints but rather on providing the final answers.

\begin{figure*}[]
  \centering
  \includegraphics[width=0.9\linewidth]{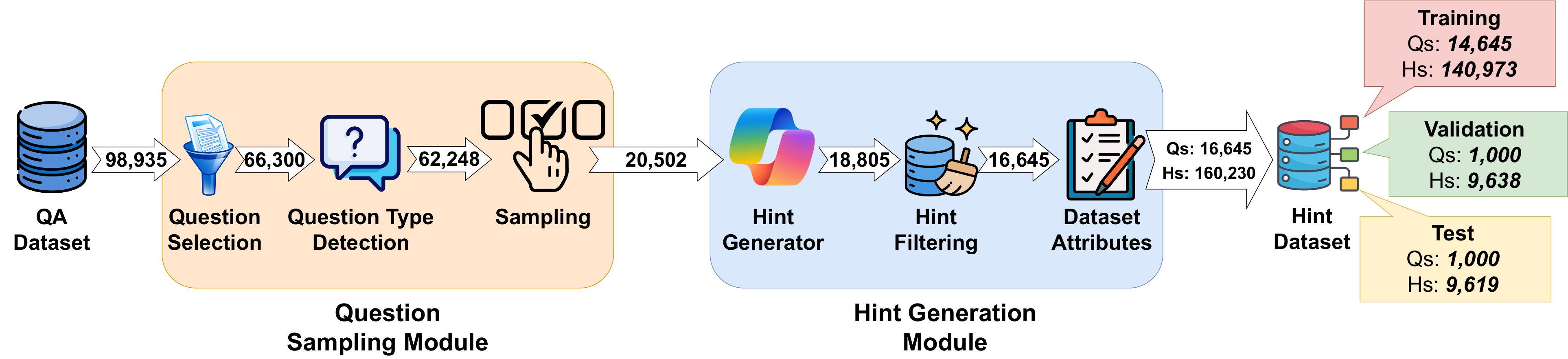}
  \caption{Dataset generation framework: Arrows represent the output quantity based on the number of questions, and callouts illustrate the statistics of the training, validation, and test sets. \textit{Qs} and \textit{Hs} denote the number of questions and hints, respectively.}
  \label{fig:dataset_framework}
\end{figure*}

\subsection{Hint Generation}\label{ss:related_hint_generation}
There are some datasets for the AHGP task such as iSnap Hint Rating\footnote{\url{https://pslcdatashop.web.cmu.edu/Project?id=321}}, and ITAP Hint Rating\footnote{\url{https://pslcdatashop.web.cmu.edu/Project?id=294}}. Both datasets comprise log data obtained from students engaged in various programming tasks, encompassing comprehensive traces of their code and logs of hint requests. The iSnap dataset~\cite{Price2019} was compiled from an introductory programming course for non-majors conducted during the Fall 2016, Spring 2017, and Fall 2017 semesters, involving 171 students who completed six programming problems. The ITAP dataset was sourced from two programming courses in Spring 2016, encompassing 89 students who tackled up to 40 Python problems~\cite{10.1145/2960310.2960333}.

The AHGP methods can be broadly classified into two categories: Data-Driven and Automated Testing-Based. Data-driven methods leverage educational data mining and machine learning techniques to automate the creation of a domain model and hints from student problem-solving data~\cite{10.1007/978-3-642-30950-2_40}. An example of this is the ITAP system, which consists of three stages: canonicalization, path construction, and reification~\cite{Rivers2017}. The Automated Testing-Based methods use automated tests to guide the hint generation process~\cite{10.1145/3430665.3456344}. For instance, the Catnip system selects suitable candidate solutions from a pool of solutions using an automated Whisker test suite~\cite{10.1145/3430665.3456344}.

Generally, existing AHGP systems generate hints only for programming tasks. In other words, none of these methods generates hints for factoid questions. To the best of our knowledge,~\citet{10.1145/3578337.3605119} is the only study focused on the AHGQ task; however, they focused on Wikipedia to generate hints without employing any LLMs. Additionally, the authors do not generate any datasets for AHGQ.

\subsection{Evaluation Methods}\label{ss:related_evaluation_methods}
There are some evaluation methods in QA, such as ExactMatch (EM), BERTScore~\cite{Zhang2020BERTScore}, BEM~\cite{bulian-etal-2022-tomayto}, and InstructGPT-eval~\cite{kamalloo-etal-2023-evaluating}. EM is an evaluation metric that measures the percentage of predicted answers that match exactly with the reference answers, providing a binary assessment of correctness. BertScore~\cite{Zhang2020BERTScore} is an evaluation metric that utilizes pre-trained BERT embeddings to measure the similarity between a generated text and a reference text. BEM~\cite{bulian-etal-2022-tomayto} is an evaluation metric that employs a fine-tuned BERT language model trained on a dataset comprising candidate answers that are equivalent to each reference answer to measure the similarity between predicted answers and reference answers. These candidate answers are generated by various QA systems trained on SQuAD~\cite{1360574093551088000}. InstructGPT-eval~\cite{kamalloo-etal-2023-evaluating} utilizes LLMs and prompts to measure the similarity of predicted answers and reference answers.

Evaluation methods for AHGP can be broadly classified into four categories: (1) User Studies and Surveys, (2) Comparisons with Experts, (3) Analysis Using Historical Data, and (4) Technical Evaluations~\cite{10.1145/3469885}. The User Studies and Surveys category involves collecting feedback from users (typically students) who interact with the hinting system~\cite{10.1145/3051457.3051467, https://doi.org/10.1111/jcal.12238}. The Comparisons with Experts category involves comparing the hints generated by the system with those provided by human experts~\cite{10.1145/2724660.2724668, price2017evaluation}. The Analysis Using Historical Data involves using past student interaction data to evaluate the effectiveness of hints~\cite{10.1007/978-3-319-61425-0_14, 10.1145/3159450.3159502}. The Technical Evaluations category involves assessing the system’s performance in terms of computational efficiency, scalability, and robustness~\cite{7930065}.

All the evaluation methods for QA are automatic, but there are a few automatic evaluation methods for AHGP. It shows that assessing hint quality automatically is a complex task. However, the automatic evaluation methods for AGHP cannot be employed to assess AGHQ because they just focus on the programming aspects of hints. As said,~\citet{10.1145/3578337.3605119} is the only study focused on the AGHQ, but they also evaluate the quality of hints manually by asking humans to assess the hints and do not propose any automatic evaluation methods for AGHQ. In our work, we introduce a fully automated method for evaluating hint qualities.

\section{Dataset}\label{s:dataset}
In this section, we outline the framework and its stages for generating the TriviaHG dataset. The framework comprises two modules: the Question Sampling Module and the Hint Generation Module. Figure~\ref{fig:dataset_framework} provides a visual representation of the framework and its stages, which we will discuss in detail in the subsequent sections.

\subsection{Question Sampling Module}\label{ss:question_sampling}
This module's task is to select questions for hint generation.
\subsubsection{Question Selection}\label{sss:question_selection}
We select the TriviaQA dataset~\cite{joshi-etal-2017-triviaqa} as our primary question source because it offers challenging trivia-style questions covering a wide range of diverse topics. It has also been frequently used in QA research \cite{beltagy2020longformer, NEURIPS2020_c8512d14, SU2024127063, prasad-etal-2023-meetingqa, Nassiri2023}. We filter out questions with fewer than 6 words or exceeding 20 words, as well as those lacking question marks. Additionally, questions whose answers do not have corresponding Wikipedia pages are excluded, as Wikipedia is utilized for evaluating hint familiarity quality. After applying these criteria, 66,300 questions remain available for further processing.

\subsubsection{Question Type Detection}\label{sss:question_type_detection}
A question classifier assigns natural language questions to classes that reflect their intended purposes. For instance, \textit{Who is Barack Obama?} falls into the \textit{PERSON} class, while \textit{Where is Washington?} belongs to the \textit{LOCATION} class. We finetune RoBERTa~\cite{liu2019roberta} model using the TREC Question Classification dataset~\cite{li-roth-2002-learning} to develop a model named QT Detector as a question classifier\footnote{The accuracy of the QT Detector on TREC Question Classification dataset is $92.8\%$}. Using the QT Detector, questions are categorized into classes at various levels of granularity, comprising 5 coarse-grained classes and 50 fine-grained classes. Subsequently, we exclude questions categorized as \textit{DESCRIPTION} type, focusing solely on factoid questions, resulting in 62,248 questions.

\subsubsection{Sampling}\label{sss:sampling}
Using the stratified sampling method~\cite{ARNAB2017213}, we conduct sampling from the questions based on their coarse-grained classes. Questions are categorized into subgroups according to their question type, and random sampling is carried out within each subgroup. In total, 20,502 questions (one-third) are sampled.

\subsection{Hint Generation Module}\label{ss:hint_generation}
This module generates hints for the selected subset of questions.

\subsubsection{Hint Generator}\label{sss:hint_generator}
\begin{figure}[]
  \centering
  \includegraphics[width=0.8\linewidth]{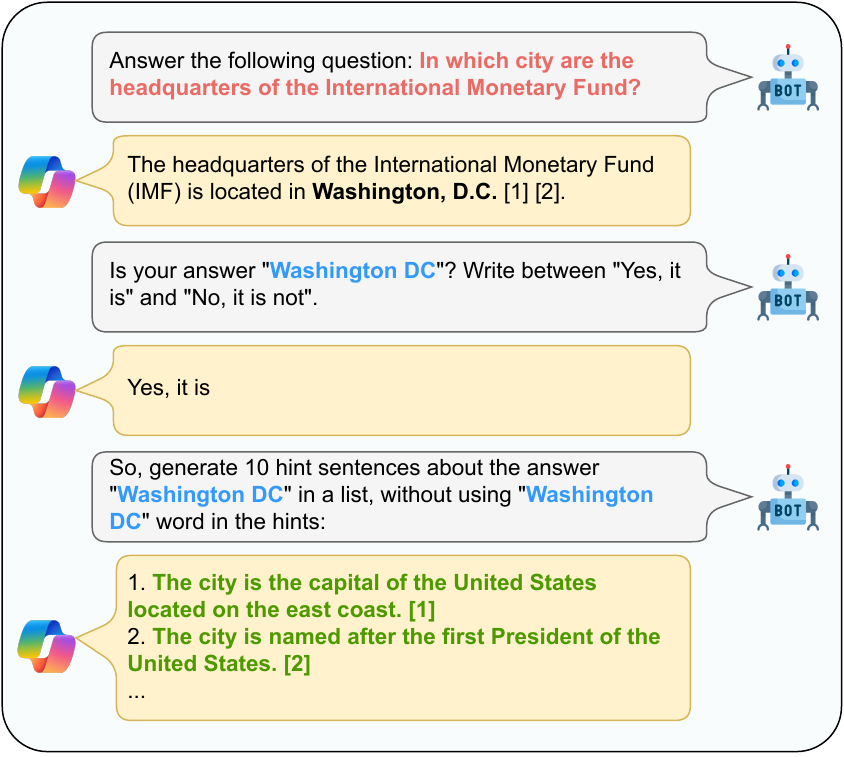}
  \caption{The hint generation system initiates by prompting \textcolor{brown}{the question} to Copilot. Then, it produces a snippet serving as the answer. Following this, we assess the correctness of the \textcolor{blue}{provided answer}. If the answer is correct, we prompt Copilot to generate 10 \textcolor{red!13!green!70!blue}{hints}. The numbers in the brackets denote the source pages for both the answer and the hints.}
  \label{fig:bing_prompting}
\end{figure}

We utilize LLMs to generate hints for the questions, specifically selecting Copilot as our preferred choice. Copilot functions as a RAG system, leveraging also information from retrieved documents on the Internet to formulate answers to user questions. This system does not only generate answers based on existing online documents but also provides the sources of the hints and answers. In contrast, LLMs like GPT-4~\cite{achiam2023gpt}, Gemini~\cite{team2023gemini}, and WizardLM~\cite{xu2023wizardlm} rely on their stored knowledge to generate answers without indicating the sources. Therefore, Copilot is considered more reliable for hint generation as it transparently presents the sources of the generated hints. To generate hints for questions, we follow these steps:

\begin{enumerate}
\item We first prompt Copilot to answer a question. This enables us to determine if Copilot is capable of retrieving the answer. If it is unable to do so, we discard the question.

\item We validate the accuracy of the generated answer by comparing it with the ground-truth answer. If they match, we proceed to the next step; otherwise, we exclude the question. This process enables us to eliminate questions whose answers may have changed, given that the TriviaQA dataset~\cite{joshi-etal-2017-triviaqa} dates back to 2017, and it is conceivable that the answers to these questions may have evolved since then. After removing such questions, 18,805 questions remained.

\item We instruct Copilot to produce 10 hints for the question without including the answer in the hints. This approach allows us to generate hints without the risk of answer leakage, i.e., the case when a hint explicitly reveals the answer to the question in its content\footnote{For instance, for the question \textit{Who won the FIVB Volleyball Men's World Championship in 2018?}, the hint \textit{Their official language is Polish.} exhibits answer leakage.}. Figure~\ref{fig:bing_prompting} illustrates the trend of using prompts in Copilot for hint generation.
\end{enumerate}

\subsubsection{Hint Filtering}\label{sss:hint_filtering}
While we advise Copilot to exercise caution regarding the problem of answer leakage, there are instances where it might produce hints inadvertently revealing the answer. Additionally, there are occasions when it generates hints that merely rephrase or describe the questions. For instance, for the question \textit{In which city are the headquarters of the International Monetary Fund?}, the model may generate a hint like \textit{The headquarters of the International Monetary Fund is located in this city}. Such hints do not help users in finding correct answers. To address these issues, we take the following steps:

\begin{enumerate}
\item For identifying potential answer leakages in the generated hints, we utilize lexical comparisons between the hints and the answer. In this process, we first tokenize the hint sentence based on words and extract the lemma of each word using the WordNetLemmatizer\footnote{We use NLTK (Natural Language Toolkit) library for word tokenizing, stop words, and lemmatization.}. Similarly, we perform the same procedure for the answer, but additionally, we exclude common stop words like \textit{a} and \textit{the} from the answer. Subsequently, if there are matching word lemmas between the hint and the answer, we eliminate the corresponding hint.

\item We utilize the SentenceTransformers framework~\cite{reimers-2019-sentence-bert} for identifying similar hints to the questions. By calculating the cosine similarity between question and hint embeddings, we establish a threshold at $0.72$\footnote{We arrived at this value by experimenting with different values.}; hints surpassing this threshold are then eliminated.

\end{enumerate}

We then eliminate questions with fewer than five hints, as we believe that having at least five hints is necessary for being able to assist users in finding the answers. By following these exclusions, 16,645 questions and 160,230 hints remain.

\begin{table}[]
    \caption{The attributes of TriviaHG dataset.}
    \label{tbl:dataset_labels}
    \resizebox{\linewidth}{!}{%
\begin{tabular}{ll}
\toprule
Name                      & Definition                                                    \\
\midrule
Q\_ID                     & Question’s ID                                                 \\
Question                  & Text of the question                                          \\
Hints                     & The generated hints                                           \\
Hints\_Sources            & The sources utilized to derive the hints.              \\
Snippet                   & The generated passage as response                      \\
Snippet\_Sources          & The sources accessed to obtain the answer.             \\
ExactAnswer               & The precise answer to the question.                           \\
MajorType                 & The coarse-grained class of the question type.                \\
MinorType                 & The fine-grained class of the question type.                  \\
Candidate\_Answers         & The candidate answers generated for convergence.      \\
Q\_Popularity             & Normalized Wikipedia page views for question entities. \\
Exact\_Answer\_Popularity & Normalized Wikipedia page views for the answer.               \\
H\_Popularity             & Normalized Wikipedia page views for hints entities.    \\
Scores                    & Result of the Candidate Evaluator component.                  \\
Convergence               & The value of HICOS                       \\
Familiarity               & The value of HIFAS                      \\
\bottomrule
\end{tabular}%
}
\end{table}

\subsubsection{Dataset Attributes}\label{sss:hint_labeling}
We prepare various attributes to be included for each question in the TriviaHG dataset, as shown in Table~\ref{tbl:dataset_labels}.

\section{Automatic Evaluation Method}\label{s:evaluation_method}
We consider there are at least five quality attributes for each hint: \emph{Relevance}, \emph{Readability}, \emph{Ambiguity}, \emph{Convergence}, and \emph{Familiarity}. Relevance evaluates how well the hint corresponds to the question, while Readability gauges its clarity and comprehensibility. Ambiguity quantifies the degree of uncertainty conveyed by the hint. Convergence quality signifies the hint's capacity to refine or exclude potential answers, while Familiarity quality measures the recognition level of entities mentioned in the hint.

In this work, we focus on automating the evaluation of convergence and familiarity quality attributes, as, in our opinion, they are the strongest indicators of hint quality. The evaluation of relevance can be conducted using, for example, the SentenceTransformers framework~\cite{reimers-2019-sentence-bert}. Additionally, to evaluate hints for their readability, factors such as hint length and the difficulty of words can be considered as well as existing readability indexes or machine learning solutions could be harnessed \cite{martinc2021supervised}. Finally, assessing the ambiguity attribute proves challenging, and finding a suitable evaluation method may be difficult. Relevance and readability will be the focus of our future work.

\subsection{Convergence Quality Attribute}\label{ss:convergence_attribute}
We are inspired by the \textit{elimination method} from cognitive science. The elimination method involves a systematic process of understanding, planning, and determining the correct answer by progressively excluding unlikely or irrelevant options based on the provided context or prior knowledge~\cite{de1996levels, tong2023eliminating}. Here, we regard the hints as the context and the stored knowledge in LLMs as prior knowledge. Subsequently, we eliminate irrelevant options\footnote{In this study, we consider candidate answers as options.} by utilizing both the provided context and prior knowledge. 

\begin{figure}[]
  \centering
  \includegraphics[width=\linewidth]{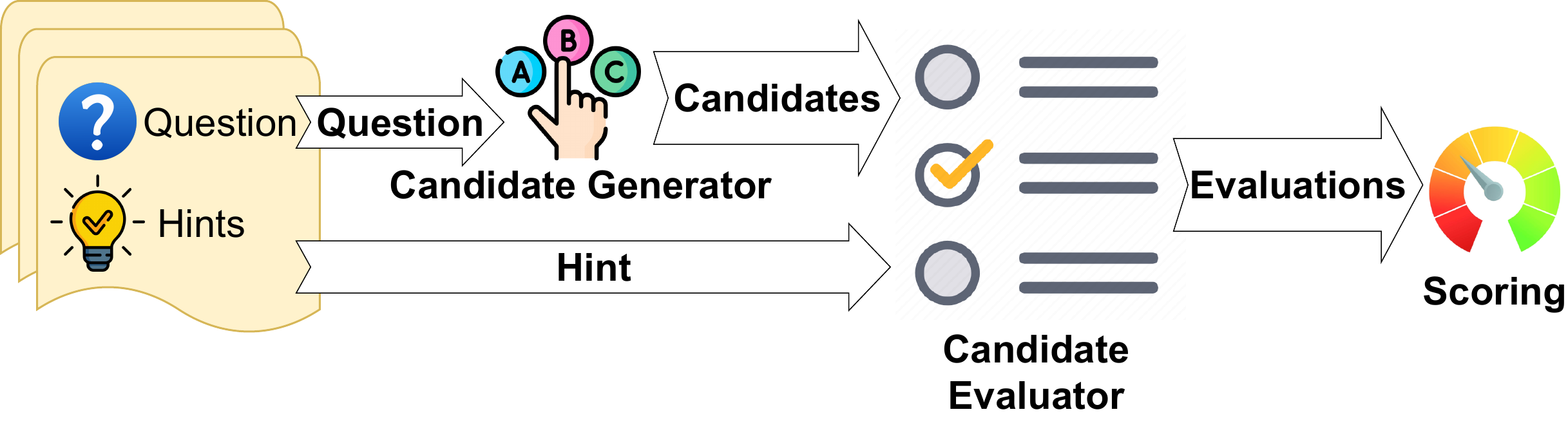}
  \caption{Convergence Evaluator: The process begins by directing the question to the Candidate Generator stage, which generates up to twenty candidate answers. Subsequently, each generated candidate answer and the hint undergo an evaluation to determine the validity of the hint for the respective candidate answer. Finally, the results are conveyed to the Scoring stage for computation of the HICOS.}
  \label{fig:convergence_evaluator}
\end{figure}

The method involves three modules: Candidate Generator, Candidate Evaluator, and Scoring. These stages are elaborated in subsequent sections, with Figure~\ref{fig:convergence_evaluator} illustrating the method's workflow.

\subsubsection{Candidate Generator}\label{sss:candidate_generator}
We employ LLMs to generate candidate answers for a given question, leveraging the vast knowledge stored within their parameters to retrieve the most potential answers. In this process, we initially prompt the LLM to generate up to $20$ candidate answers for the given question.\footnote{For example, when considering the question \textit{In which city are the headquarters of the International Monetary Fund?}, some of the generated candidate answers include \textit{Brussels, Vienna, Paris, Tokyo, Moscow, Stockholm, Amsterdam, London, Edinburgh, etc.}} The use of \textit{up to} in the prompt accounts for instances where questions may not have $20$ candidate answers. For instance, in the question \textit{If Christmas Day falls on a Tuesday, on what day will New Year's Day fall?} with the answer \textit{Tuesday}, there are only seven candidate answers since they correspond to weekdays. The prompt we employed is as follows:

\begin{verbatim}
  Generate up to 20 candidate answers words for the quest
  ion "QUESTION" in bullet points.
\end{verbatim}

In the prompt, the term \textit{QUESTION} pertains to the provided question. Then, we analyze the output, extract the candidate answers from the list, and feed them along with the hint into the Candidate Evaluator module.

\subsubsection{Candidate Evaluator}\label{sss:candidate_generator}
We employ LLMs to determine if the hint pertains to the generated candidate answers. This is accomplished by prompting the model with the following prompt:

\begin{verbatim}
  Does the hint "HINT" refer to "CANDIDATE"? Choose ONLY
  between "Yes" or "No".
\end{verbatim}

In the prompt, the term \textit{HINT} corresponds to the provided hint, while \textit{CANDIDATE} corresponds to one of the generated candidate answers. This prompt is utilized for all the generated candidate answers, and the resulting outputs, which include lists of \textit{Yes} and \textit{No}, are directed to the Scoring module. 
We interpret \textit{Yes} as one and \textit{No} as zero.

\subsubsection{Scoring}\label{sss:scoring}
In this module, we utilize the following equation to calculate the HICOS of the hint. To assess the convergence quality, if the Candidate Evaluator module identifies that the hint is not valid for the answer, we assign a score of zero, indicating that the hint is not useful for finding the answer. However, if the hint is valid for the answer, we compute the score based on the results for other candidate answers. A score of zero denotes that the hint does not reduce the number of candidate answers, whereas a score of one indicates that the hint successfully eliminates all candidate answers except for the correct one.

\begin{equation}
    Score_{con}= \left\{ \begin{array}{ll}
    0 & : \ EA_{valid} = 0 \\
    1-\frac{(\sum_{cand} cand_{valid})-1}{\left| candidates \right|} & : \ EA_{valid} = 1 \\
    \end{array} \right.
\end{equation}

In the given equation, \textit{cand} and $cand_{valid}$ denote the candidate answer and the outcome of the Candidate Generator module for it, respectively, and $|candidates|$ represents the total number of candidate answers generated by the Candidate Generator module. $EA_{valid}$ equals one if the ground-truth answer is predicted as a valid candidate answer; otherwise, it is zero.

\subsection{Familiarity Quality Attribute}\label{ss:familiarity_attribute}
We aim to measure the familiarity of information expressed in hints (HIFAS). Here, we use the popularity of the mentioned entities as a proxy for familiarity. The presence of popular entities can assist users in locating the answer more effortlessly, while less popular entities in a hint may potentially lead the user astray in finding the correct answer~\cite{10.1007/978-3-030-75015-2_8}. 

\begin{figure}[]
  \centering
  \includegraphics[width=\linewidth]{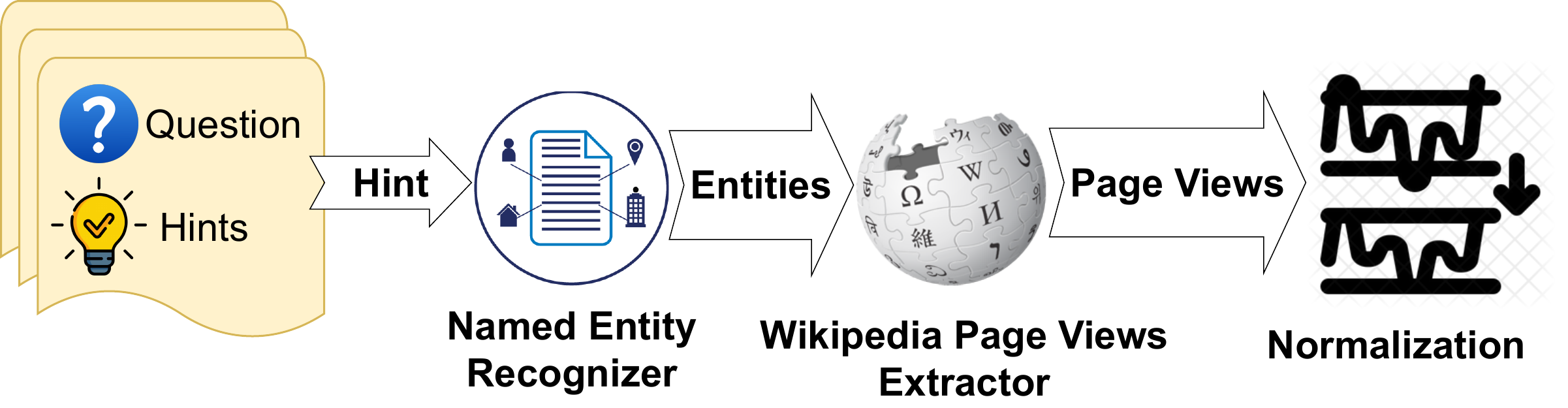}
  \caption{Familiarity Evaluator: The Named Entity Recognizer identifies named entities from the hint. Subsequently, the number of views for the Wikipedia page associated with each entity is extracted. Finally, the view count for each entity undergoes a normalization process.}
  \label{fig:familiarity_evaluator}
\end{figure}

The method involves three steps: Named Entity Recognition (NER), Wikipedia Page Views Extractor, and Normalization. These steps are illustrated in Figure~\ref{fig:familiarity_evaluator}.

We utilize the spaCy library\footnote{\url{https://www.spacy.io/}} for extracting named entities from the hints. Subsequently, we leverage the Pageview API\footnote{We use the pageview-api library available at \url{https://github.com/Commonists/pageview-api}} to retrieve the number of views for the Wikipedia pages associated with the extracted entities, covering the period from January 1, 2015, to December 31, 2023, and we compute the number of views for each month. Due to the high variability in the number of page views, we normalize them to a range between zero and one. 

It is worth noting that some entities exhibit extremely high (or extremely low) numbers of views\footnote{For example, the entity \textit{India} garners a total of 2,685,795 page views per month.}. When attempting to normalize the number of page views, such entities can render the normalization values unreliable due to their unusual number of views. To address this issue, we collect view counts for approximately 50,000 different entities. Subsequently, we utilize the Interquartile Range (IQR) method to identify outliers in the view counts. Defining the IQR as the difference between the $Q_3$ (the 75th percentile) and $Q_1$ (the 25th percentile), data points falling below $min=Q_1 - 1.5 \times IQR$ or exceeding $max=Q_3 + 1.5 \times IQR$ are classified as outliers~\cite{whaley2005interquartile}. Following this classification, we truncate values exceeding $max$ to $max$ and values falling below $min$ to $min$. By implementing this approach, the normalization process yields more reliable results. Finally, we employ linear scaling~\cite{han2022data} to normalize the number of views, bringing them into a range between 0 and 1.

\begin{figure}[]
  \centering
  \includegraphics[width=0.9\linewidth]{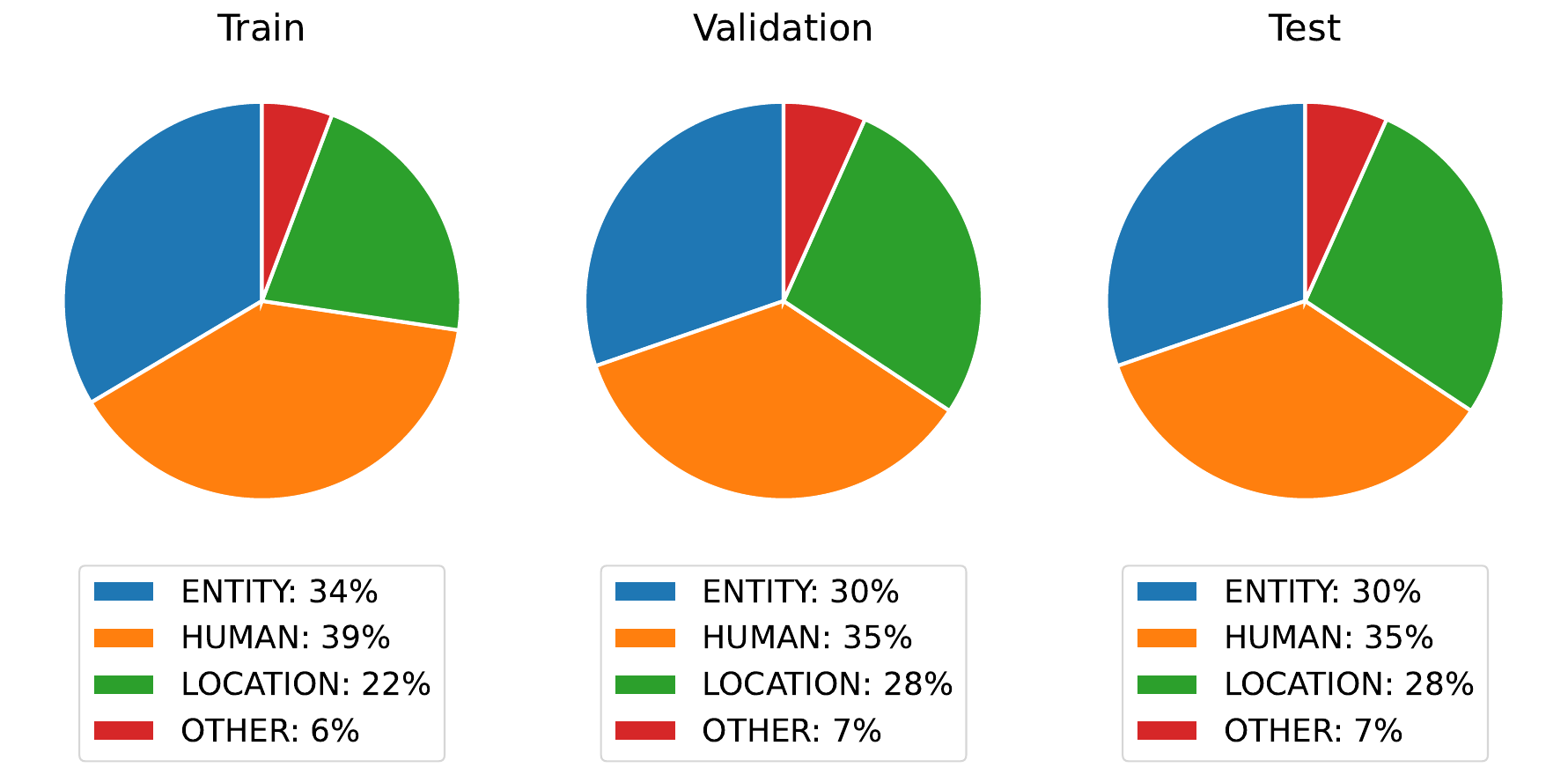}
  \caption{The distribution of Train, Validation, and Test set.}
  \label{fig:sub_distributions}
\end{figure}

\section{Experiments and Results}\label{s:experiments}
In this section, we first explore dataset features, including statistics, difficulty, and quality. Subsequently, we assess our automatic evaluation method, and finally, we investigate the performance of different models on the TriviaHG dataset using the proposed automatic evaluation method.

\begin{table}[]
    \small
    \caption{Statistics of the TriviaHG dataset}
    \label{tbl:statistics}
    \begin{tabular}{@{}p{0.24\textwidth}lll@{}}
        \toprule
                                     & Training  & Validation  & Test \\ 
        \midrule
        Number of questions          & 14,645  & 1,000   & 1,000   \\
        Number of hints              & 140,973 & 9,638   & 9,619   \\
        \midrule
        Avg. question length (words) & 14.18 & 14.08 & 13.95 \\
        Avg. hint length (words)     & 14.98 & 15.07 & 15.14 \\
        Avg. \#hints / question        & 9.62  & 9.63  & 9.61  \\
        Avg. \#entities / question     & 1.35  & 1.40  & 1.35  \\
        Avg. \#entities / hint         & 0.96  & 1.00  & 0.98  \\
        Avg. \#sources / question      & 6.27  & 6.17  & 6.71  \\
        \bottomrule
    \end{tabular}
\end{table}

\subsection{Data Analysis}\label{ss:data_analysis}
\subsubsection{Dataset Statistics}\label{sss:data_statistics}
Employing the framework shown in Figure~\ref{fig:dataset_framework}, we generated a dataset comprising 16,645 questions and 160,230 hints extracted from the TriviaQA~\cite{joshi-etal-2017-triviaqa} dataset. The attributes associated with questions are detailed in Table~\ref{tbl:dataset_labels}. TriviaHG is randomly divided into a training set, consisting of 140,973 hints (14,645 questions), a validation set with 9,638 hints (1,000 questions), and a test set comprising 9,619 hints (1,000 questions). Table~\ref{tbl:statistics} provides an overview of the dataset statistics. The numbers indicate that the characteristics of the sub-datasets bear a close resemblance to each other.

We also conducted an analysis of sub-datasets categorized by the coarse-grained class of the question type. Illustrated in Figure~\ref{fig:sub_distributions}, the sub-datasets exhibit a close distribution, with the highest number of questions focused on HUMAN, followed by ENTITY, LOCATION, and lastly, OTHER\footnote{The \textit{OTHER} category encompasses questions related to numbers, dates, etc.}.

\subsubsection{Difficulty}\label{sss:difficulty}
\begin{figure}[]
  \centering
  \includegraphics[width=0.85\linewidth]{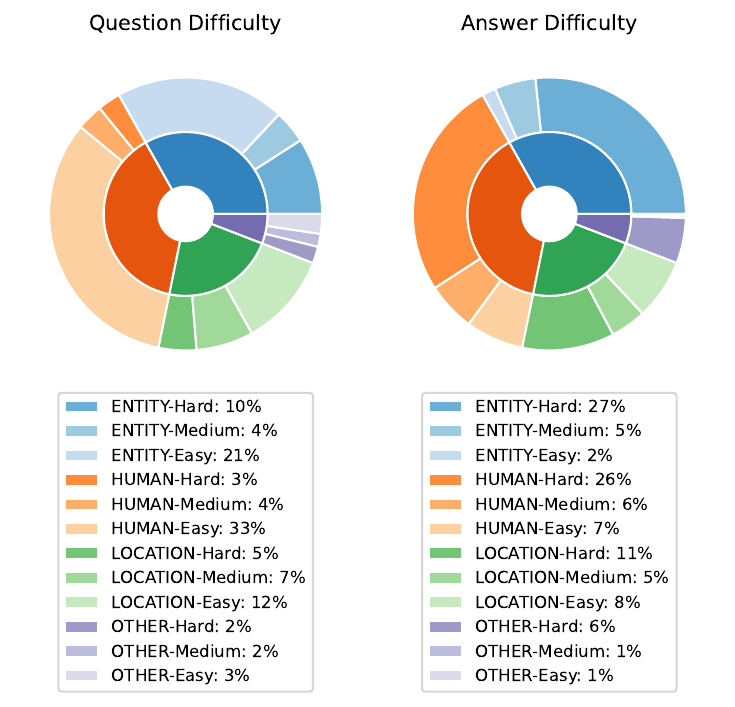}
  \caption{The Left figure illustrates question difficulty, while the Right figure presents answer difficulty.}
  \label{fig:difficulty}
\end{figure}

We conduct a comprehensive examination of the dataset concerning the difficulty levels of both questions and answers. To evaluate the difficulty of questions, we utilize the DPR method~\cite{karpukhin-etal-2020-dense} with the assistance of the Pyserini toolkit~\cite{10.1145/3404835.3463238} to retrieve 500 relevant passages from the English Wikipedia that correspond to the questions. If the relevance of passages to a question is below $33\%$, we classify it as hard; if it is below $66\%$, we classify it as medium; and if it is more than $66\%$, we consider it easy. Regarding the difficulty assessment for answers, we adopt the methodology outlined in Section~\ref{ss:familiarity_attribute} to quantify and normalize the popularity of named entities of answers. If the popularity score exceeds 0.66, we categorize an answer as easy; if it falls within the range of 0.33 to 0.66, we label it as medium; and if the popularity is less than 0.33, we designate it as hard. Figure~\ref{fig:difficulty} shows the distribution of questions and answers based on the difficulty.

\subsection{Human Evaluation}\label{sss:human_evaluation}
We engage human evaluators to assess the quality of the generated hints. To accomplish this, we randomly selected 100 questions from the test set, distributed as 25 questions from each coarse-grained class of question types (Entity, Human, Location, and Other), consisting of 9 labeled as hard, 8 labeled as medium, and 8 labeled as easy. In addition to the TriviaHG hints, we instruct LLaMA\footnote{Please note that we employ LLaMA version 2 for this study.} 7b~\cite{touvron2023llama} vanilla to generate $10$ hints for each question\footnote{We utilize the identical prompt as in Copilot for generating hints.}. Furthermore, we finetune another LLaMA 7b instance on the training set of TriviaHG and then prompt it to generate $10$ hints for each question\footnote{We perform model finetuning using the API functions available on AnyScale.com}. Subsequently, we employ the hint filtering stage (Section \ref{sss:hint_filtering}) to refine the generated hints. At last, we obtain a total of 2,741 hints, comprising 950 hints from the TriviaHG dataset, 901 hints from the finetuned LLaMA 7b, and 890 hints from the LLaMA 7b vanilla. These data undergo evaluation in two distinct phases:

In the first phase, we ask $10$ human evaluators to evaluate the generated hints, providing ratings for attributes such as relevance, readability, ambiguity, convergence, and familiarity. Ratings are assigned on a scale of 1 to 5, where 1 denotes the lowest and 5 signifies the highest. They are also asked to search for each hint on both Google and Bing search engines, quickly examine the result page within a 5-second timeframe, and then determine and declare whether they could find the answer on the search results page. We request this action as it also can signal the quality of the hints. The reason is that it is plausible that a hint is still effective, while the knowledge of the annotators might be insufficient to answer the question utilizing the hint. The results of the human evaluation are summarized in Table~\ref{tbl:human_evaluation_attributes}, displaying average scores for each quality attribute. The relevance analysis reveals that hints generated by Copilot and the finetuned LLaMA 7b are relevant to the questions, while those from LLaMA 7b vanilla lack in terms of relevance. Additionally, the assessments of readability and ambiguity indicate that the generated hints are clear and comprehensible. The analysis of convergence and familiarity indicates that Copilot exhibits higher convergence and familiarity than other methods, while the finetuned LLaMA 7b demonstrates superior quality compared to the LLaMA 7b vanilla. Furthermore, Table~\ref{tbl:search_engines} provides statistics on hints that annotators could use to find answers to the respective questions through searches on Google and Bing search engines.

\begin{table}[]
    \caption{The human evaluation results for quality attributes.}
    \label{tbl:human_evaluation_attributes}
    \resizebox{\columnwidth}{!}{%
    \begin{tabular}{@{}llllll@{}}
    \toprule
    Method            & Relevance & Readability & Ambiguity & Convergence & Familiarity \\
    \midrule
    Copilot              & 4.09      & 4.67        & 1.51      & 2.23        & 2.47        \\
    LLaMA - Finetuned & 4.01      & 4.7         & 1.56      & 2.20        & 2.41        \\
    LLaMA - Vanilla   & 3.64      & 4.47        & 1.87      & 2.12        & 2.02         \\
    \bottomrule
    \end{tabular}%
    }
\end{table}

\begin{table}[]
    \small
    \caption{The count of hints that accurately reveal the correct answer when evaluators search for them on search engines.}
    \label{tbl:search_engines}
    \begin{tabular}{@{}p{0.2\textwidth}lll@{}}
        \toprule
        Method            & Num. of Hints & Google     & Bing       \\
        \midrule
        Copilot             & 950           & 734 (77\%) & 707 (74\%) \\
        LLaMA - Finetuned & 901           & 372 (41\%) & 353 (39\%) \\
        LLaMA - Vanilla   & 890           & 228 (26\%) & 234 (26\%) \\
        \bottomrule
    \end{tabular}
\end{table}

\begin{figure*}[]
  \centering
  \includegraphics[width=0.8\linewidth]{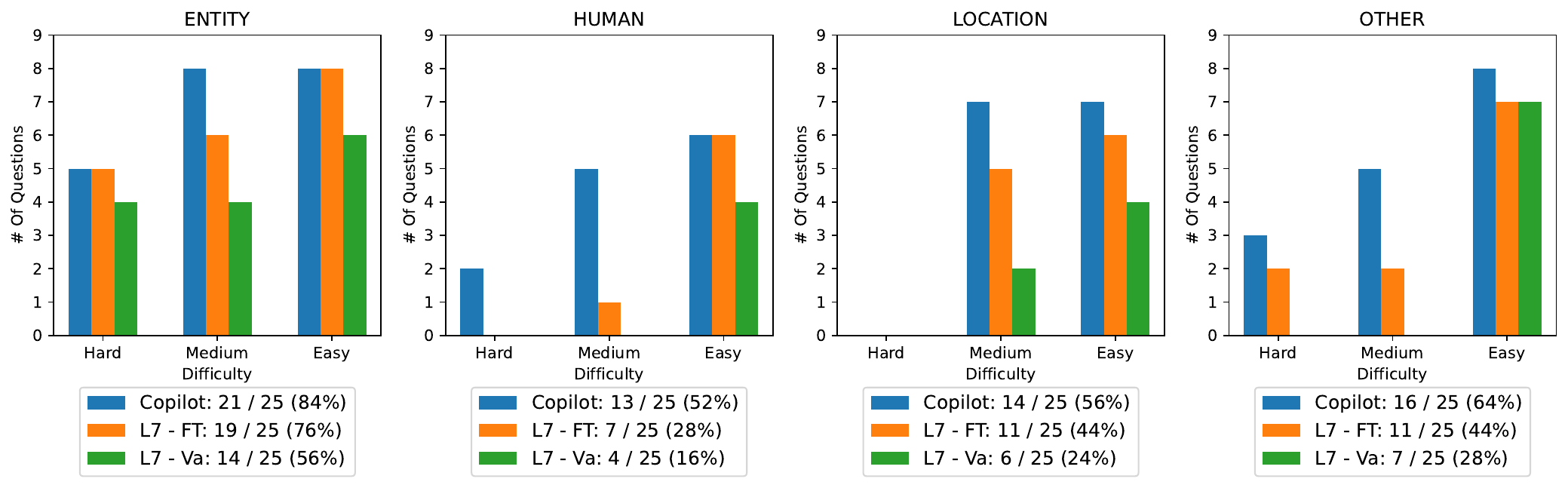}
  \caption{The statistics of human evaluation. \textit{L7 - Va} refers to \textit{LLaMA 7b Vanilla} and \textit{L7 - FT} refers to \textit{LLaMA 7b Finetuned}.}
  \label{fig:human_evaluation_questions}
\end{figure*}

In the second phase, six distinct individuals, entirely separate from those involved in the first phase, are tasked with answering the questions using the provided hints in a similar way to the approach in \cite{10.1145/3578337.3605119}. We randomly select hints for each question from Copilot, LLaMA 7b vanilla, and finetuned LLaMA 7b. The procedure is as follows:

\begin{enumerate}
\item The subjects are asked to answer the question without employing any hints. If they can answer the question, they can go to the next question.
\item If they cannot answer the question, the annotators are asked to read hints one by one until they come up with the answer. If their answer is correct, they can move to the next question.
\item If participants cannot answer the question by reading all of the hints of the question, they can leave the question.
\end{enumerate}

Figure~\ref{fig:human_evaluation_questions} provides an overview of the questions successfully answered by participants. The data indicates the high effectiveness of hints, as participants could answer only 19 questions before consulting hints, whereas after reviewing the hints, they could successfully answer 67 questions. The results from human evaluations underscore the superior efficacy of Copilot's hints in aiding users in formulating answers compared to other methods. Our findings show that finetuned LLaMA 7b hints offer more value than LLaMA 7b vanilla hints, suggesting that leveraging the TriviaHG dataset for LLM finetuning contributes to the generation of superior hints. Only $36\%$ of questions with hard answers were answerable, while medium and easy answers boasted success rates of $78\%$ and $96\%$, respectively. This observation emphasizes the increased challenge in generating hints for questions with difficult answers compared to their easier counterparts. Moreover, Figure~\ref{fig:human_evaluation_questions} highlights that $64\%$ of the questions could be answered using Copilot hints, while the respective figures for the finetuned LLaMA 7b and LLaMA 7b vanilla are $48\%$ and $31\%$. It demonstrates the quality disparity between Copilot and the finetuned and vanilla LLaMA 7b hints.

\subsection{Automatic Evaluation Method}\label{ss:evaluation_method}
We assess the automatic evaluation method by comparing the human-provided scores with the generated HICOS and HIFAS.

\begin{figure}[]
  \centering
  \includegraphics[width=0.9\linewidth]{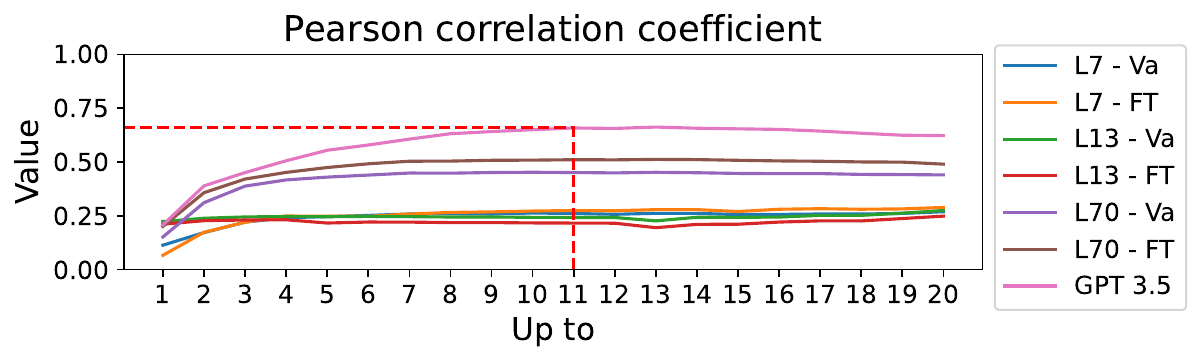}
  \caption{The Pearson Correlation Coefficient between human assessments and value of HICOS. \textit{L7, L13, L70} refers to \textit{LLaMA 7b, 13b, 70b}, respectively.}
  \label{fig:pearson_mse_convergence}
\end{figure}

\subsubsection{Convergence Quality Attribute}\label{sss:convergnce_quality}
As elaborated in Section~\ref{ss:convergence_attribute}, we leverage LLMs as the kernel of the automatic evaluation method to assess the convergence quality attribute of a hint (HICOS). To this end, we employ LLaMA 7b, LLaMA 13b, LLaMA 70b, GPT 3.5~\cite{brown2020language} models, as well as the versions of LLaMA 7b, LLaMA 13b, and LLaMA 70b finetuned on the training set of TriviaHG to examine the accuracy of the evaluation method across different kernels. In addition, as discussed in Section~\ref{sss:scoring}, the value of HICOS is influenced by the number of candidate answers generated. To determine the optimal model as the kernel and the optimal number of candidate answers, we calculate the Pearson Correlation Coefficient~\cite{han2022data} between human assessments and the value of HICOS across 2,791 hints. Figure~\ref{fig:pearson_mse_convergence} indicates the value of the Pearson Correlation Coefficient for each number of candidate answers for the various kernels. Based on the observations, we can infer that the evaluation method's alignment with human assessments for the convergence quality attribute is closely linked to the robustness of the LLM used as the kernel. Notably, the highest correlation is observed with GPT 3.5, followed by finetuned LLaMA 70b and LLaMA 70b vanilla. Furthermore, the findings suggest that finetuned LLaMA on the training set yields evaluations more akin to human judgments compared to LLaMA vanilla models. Moreover, the values indicate that generating $11$ candidate answers yields the closest correlation to human results.

\subsubsection{Familiarity Quality Attribute}\label{sss:familiarity_quality}
As detailed in Section~\ref{ss:familiarity_attribute}, we employ the normalized number of page views of the Wikipedia page corresponding to entities as the measure of familiarity quality attribute for hints (HIFAS). For each hint, we explore three distinct methods, including the minimum, average, and maximum. In the minimum (maximum) method, we solely consider the normalized value of an entity that has the minimum (maximum) number of views. Conversely, in the average method, we compute the average of normalized values across all the entities mentioned in the hint. To determine the best approach, we compute the Pearson Correlation Coefficient and Mean Squared Error (MSE)~\cite{han2022data} between human assessments and the value of HIFAS across 2,791 hints. Table~\ref{tbl:pearson_mse_familiarity} indicates the value of the Pearson Correlation Coefficient and Mean Squared Error for all the above approaches. The findings suggest that calculating the average of normalized page view values across entities exhibits the highest correlation with human evaluations.

\begin{table}[]
    \small
    \caption{The Pearson Correlation Coefficient and MSE between human assessments and HIFAS across 2,791 hints.}
    \label{tbl:pearson_mse_familiarity}
    \begin{tabular}{@{}p{0.23\textwidth}lll@{}}
        \toprule
        Method   & Minimum & Average & Maximum \\
        \midrule
        Pearson Correlation Coefficient & 0.63     & \textbf{0.66}    & 0.63    \\
         Mean Squared Error     & 0.16    & \textbf{0.13}    & 0.14   \\
        \bottomrule
    \end{tabular}
\end{table}

\subsection{Model Performance}\label{ss:model_performance}
Finally, we utilize LLaMA, GPT 3.5, WizardLM 70b, Gemini, GPT 4, and the finetuned versions of LLaMA on the training set of TriviaHG to generate hints, followed by evaluation using the automatic evaluation method. For evaluating the convergence quality attribute, we generate $11$ candidate answers and utilize finetuned LLaMA 70b as the kernel model\footnote{The decision to choose the second-best model is driven by the high cost of using GPT 3.5 and its closed-source character.}. Meanwhile, in evaluating the familiarity quality attribute, we employ the average approach. Table~\ref{tbl:model_performance} presents the outcomes of the automatic evaluation method for different LLMs.

The findings suggest that a robust model, characterized by a greater number of training parameters, produces hints of higher convergence quality. This may be attributed to its capacity to store more knowledge within its parameters, resulting in more accurate hints. Furthermore, the results indicate that finetuned LLaMA models outperform LLaMA vanilla models, demonstrating that finetuning on TriviaHG enhances the hint generation capabilities of LLMs. Moreover, comparing the results between Copilot and GPT 4 underscores the effectiveness of incorporating retrieved documents alongside stored knowledge. This is exemplified by Copilot, which utilizes GPT 4 as its kernel.

\begin{table}[]
    \small
    \caption{The HICOS and HIFAS values across 100 questions for different models.}
    \label{tbl:model_performance}
        \begin{tabular}{@{}p{0.35\textwidth}ll@{}}
        \toprule
        Model        & HICOS & HIFAS    \\
        \midrule
        LLaMA 7b - Vanilla~\cite{touvron2023llama}     & 0.307       & 0.833          \\
        LLaMA 13b - Vanilla~\cite{touvron2023llama}    & 0.350       & 0.929          \\
        LLaMA 70b - Vanilla~\cite{touvron2023llama}    & 0.425       & 0.941          \\
        \midrule
        LLaMA 7b - Finetuned  & 0.400       & 0.890          \\
        LLaMA 13b - Finetuned & 0.410	    & 0.881          \\
        LLaMA 70b - Finetuned & 0.494	    & 	0.862        \\
        \midrule
        GPT 3.5~\cite{brown2020language}      & 0.438	       & 0.911          \\
        WizardLM 70b~\cite{xu2023wizardlm} & 0.446      & 0.942          \\
        Gemini~\cite{team2023gemini}       & 0.455      & 0.911          \\
        GPT 4~\cite{achiam2023gpt}        & 0.525       & 0.875          \\
        \midrule
        Copilot        & \textbf{0.540}       & \textbf{0.946}      \\
        \bottomrule
        \end{tabular}
\end{table}

\section{Limitations and Use Cases}\label{s:limitations}
Our approach and evaluation method have some limitations:
\begin{itemize}
    \item Our focus was solely on factoid questions, where the answers are named entities, each with its own Wikipedia page.
    
    \item Our dataset and evaluation method depend on LLMs; thus, the biases and limitations inherent in these models will be reflected in both our dataset and evaluation approach.
    
    \item We solely assess the model performance based on the convergence and familiarity quality attributes.

    \item In the convergence quality evaluation method (HICOS), we found 11 to be the optimal number of candidate answers on TriviaHG. However, this number could vary for other datasets. When applying the evaluation method to other datasets, one could ideally try to determine the optimal number of candidate answers for the particular used data.
    
    \item The current approach focuses solely on generating hints using LLMs and assessing their usefulness in assisting humans to answer questions without explicitly considering the educational values of hints.
\end{itemize}

Our dataset and evaluation method can be used in different ways:
\begin{itemize}
    \item The dataset holds the potential for finetuning generative models to produce high-quality hints for the provided questions. As illustrated in Figure~\ref{fig:human_evaluation_questions} and Table~\ref{tbl:model_performance}, finetuned models trained on our dataset have demonstrated improved hint generation compared to vanilla versions.
    
    \item The proposed evaluation method for assessing the convergence and familiarity quality attributes can serve as a tool to evaluate the quality of hints.
    
    \item The proposed dataset could be employed for improving QA systems by adding hints as additional context to prompts similar to~\citet{sun2023autohint}.
    
    \item The generated hints could be of use in diverse information retrieval tasks. For instance, they can be employed in retrieval-augmented generation (RAG) techniques to improve question understanding. Additionally, the hints can be utilized in query expansion methods. They could also be used in entertainment contexts, such as in game shows like Jeopardy.

\end{itemize}
\section{Conclusion}\label{s:conclusion}
Hinting is an important and frequent mechanism in human question answering; however, it has been largely neglected during the recent advancement of QA technologies. In this paper, we present a framework for constructing datasets tailored to hint generation for factoid questions, consisting of two key modules: 1) Question Sampling Module and 2) Hint Generation Module. The framework was utilized to generate a dataset comprising 140,973 hints (14,645 questions) in the training set, 9,638 hints (1,000 questions) in the validation set, and 9,619 hints (1,000 questions) in the test set.

Additionally, an automatic evaluation method is proposed to measure the convergence and familiarity quality attributes of the hints. Convergence quality is evaluated using LLMs as a kernel framework, while familiarity quality is assessed using the number of page views of Wikipedia page entities. Human assessment of the dataset and evaluation method indicates high-quality hints and a strong correlation with human results for the evaluation method. Finally, the performance of various LLMs on the dataset was evaluated based on convergence and familiarity quality attributes, demonstrating the impact of the dataset on finetuning.

In the future, our plan includes generating hints based on user feedback, integrating the subjective aspect of hints, and evaluating hints based on additional quality attributes.

%%\begin{acks}
%%To Robert, for the bagels and explaining CMYK and color spaces.
%%\end{acks}

\bibliographystyle{ACM-Reference-Format}
\balance
\bibliography{references}


\end{document}